\documentclass[11pt]{article}
\usepackage{graphicx}
\usepackage{times}
\usepackage{amsmath}
\usepackage{cite}
\usepackage{setspace}
\usepackage{amssymb}
\usepackage{algorithm}
\usepackage{algpseudocode}
\usepackage{hyperref}

\usepackage[left=1.5in, right=1.5in, top=1in, bottom=1in]{geometry}

\title{Coordinate Heart System: A Geometric Framework for Emotion Representation}
\author{Omar Aldesi \\ 
\textit{Jordan, Amman} \\
\href{mailto:2441394@std.hu.edu.jo}{\texttt{2441394@std.hu.edu.jo}}
}
\date{June 2025}

\begin{document}

\maketitle

\begin{abstract}
This paper presents the Coordinate Heart System (CHS), a geometric framework for emotion representation in artificial intelligence applications. We position eight core emotions as coordinates on a unit circle, enabling mathematical computation of complex emotional states through coordinate mixing and vector operations. Our initial five-emotion model revealed significant coverage gaps in the emotion space, leading to the development of an eight-emotion system that provides complete geometric coverage with mathematical guarantees.

The framework converts natural language input to emotion coordinates and supports real-time emotion interpolation through computational algorithms. The system introduces a re-calibrated stability parameter $S \in [0,1]$, which dynamically integrates emotional load, conflict resolution, and contextual drain factors. This stability model leverages advanced Large Language Model (LLM) interpretation of textual cues and incorporates hybrid temporal tracking mechanisms to provide nuanced assessment of psychological well-being states.

Our key contributions include: (i) mathematical proof demonstrating why five emotions are insufficient for complete geometric coverage, (ii) an eight-coordinate system that eliminates representational blind spots, (iii) novel algorithms for emotion mixing, conflict resolution, and distance calculation in emotion space, and (iv) a comprehensive computational framework for AI emotion recognition with enhanced multi-dimensional stability modeling. 

Experimental validation through case studies demonstrates the system's capability to handle emotionally conflicted states, contextual distress factors, and complex psychological scenarios that traditional categorical emotion models cannot adequately represent. This work establishes a new mathematical foundation for emotion modeling in artificial intelligence systems, with applications in human-computer interaction, mental health monitoring, and affective computing.

The mathematical framework has been fully implemented in software (Section 9) to facilitate reproducibility and practical adoption.

\end{abstract}
\section{Introduction}

Emotion recognition and modeling represent fundamental challenges in artificial intelligence, with applications that span human-computer interaction, mental health monitoring, and affective computing systems. Traditional approaches to emotion representation rely primarily on categorical models that classify emotions into discrete states, or dimensional models that position emotions along continuous axes such as valence and arousal~\cite{russell1980circumplex}. Although these frameworks have proven useful, they often lack the mathematical precision required for computational systems to accurately model the complex, continuous nature of human emotional experience.

The limitations of existing approaches become particularly apparent when attempting to represent mixed or complex emotional states. Current categorical systems struggle to capture emotions that exist between traditional categories, while dimensional models, though continuous, often fail to provide intuitive geometric relationships that correspond to human emotional experience. This gap between theoretical emotion models and practical computational requirements has motivated the development of new mathematical frameworks for emotion representation.

This paper introduces the Coordinate Heart System (CHS), a novel geometric approach to emotion modeling that addresses these limitations through a mathematically grounded coordinate-based framework. Beyond core emotions, CHS uniquely integrates a **re-calibrated Stability parameter, now operating on a [0, 1] scale**, allowing for a comprehensive representation of psychological well-being influenced by both emotional load and non-emotional contextual factors. The system positions core emotions as fixed points on a unit circle, enabling complex emotional states to emerge through vector mathematics and coordinate interpolation, while its refined stability model ensures a nuanced and realistic understanding of a user's state. Our approach provides both the precision required for computational applications and the geometric intuition that aligns with human emotional experience.

The development of CHS revealed a critical insight: Initial attempts using five core emotions resulted in significant coverage gaps within the emotion space, creating "blind spots" where certain emotional states could not be accurately represented. Through systematic analysis, we identified that eight strategically placed emotions provide complete geometric coverage of the emotional space, eliminating these representation gaps.

The primary contributions of this work include: (1) a mathematical proof demonstrating why five-emotion coordinate systems are insufficient for complete emotional space coverage, (2) the design and validation of an eight-emotion coordinate system that eliminates coverage gaps, (3) algorithms for emotion mixing, interpolation, and distance calculation within the coordinate framework, and (4) a **robust computational architecture suitable for real-time AI emotion recognition applications, featuring an enhanced stability model that dynamically incorporates contextual influences.**

Details on the software implementation and code availability are provided in Section 9.

\section{Related Work}
The field of computational emotion modeling has evolved through several distinct approaches, each attempting to bridge the gap between human emotional experience and mathematical representation.
\subsection{Categorical Emotion Models}
Early computational systems relied on Ekman's six basic emotions~\cite{ekman1992argument} and Plutchik's wheel of emotions~\cite{plutchik2001nature}. While classification-friendly, these models struggle with continuous emotional gradations and mixed states~\cite{barrett2006solving}.
\subsection{Dimensional Emotion Models}
Russell's circumplex model~\cite{russell1980circumplex} represents emotions using valence and arousal dimensions, with extensions to PAD (Pleasure-Arousal-Dominance) models~\cite{mehrabian1996pleasure}. However, these abstract dimensions lack intuitive geometric relationships and require complex calibration~\cite{posner2005circumplex}.
\subsection{Existing Coordinate Systems}
Current coordinate-based approaches focus on abstract mathematical spaces rather than interpretable geometric systems~\cite{fontaine2007world}. Graph-based models capture transitions but lack continuous coordinate precision~\cite{kim2013emotional}. Lövheim's cube model~\cite{lovheim2012new} provides geometric structure but targets neurobiological rather than computational applications.
\subsection{Research Gap}
No existing framework combines: (1) explicit 2D coordinate semantics, (2) 8-core emotion integration, (3) stability-aware interpolation, and (4) efficient computational encoding. Our work addresses this gap by providing a unified mathematical framework for precise emotional state computation and representation.

\section{Theoretical Foundation}

\subsection{Core Concept: A Geometric Metaphor for Emotion}

This research proposes a mathematical framework for representing emotions through a novel geometric metaphor. The inspiration for this model came from a conceptual exercise in mapping emotional states to a coordinate system, heuristically anchored to the idea of a 'heart center'.

This conceptual mapping led to a geometric arrangement where the emotional coordinates form a unit circle, providing a computationally convenient and intuitive structure for an AI model.
\subsection{Mathematical Formulation of Coordinate System}

The coordinate system is established with the cardiac center as the origin, where the first quadrant of the unit circle extends toward the right side of the body. The eight core emotions—anger, sadness, guilt, pride, love, disgust, joy, and fear—are mapped to specific angular positions on this unit circle.

Each emotion's spatial location was determined by connecting the observed somatic points, which approximated a unit circle centered on the cardiac region. The mathematical coordinates for each emotion are derived using standard trigonometric functions:
\begin{align}
x &= \cos(\theta) \\
y &= \sin(\theta)
\end{align}
where $\theta$ represents the angular position of each specific emotion on the unit circle.

This coordinate system serves as the foundation for representing the eight basic emotions, while complex mixed emotions require additional mathematical formulations that will be addressed in subsequent sections.
\subsection{Emotion Mixing Theory: How Complex Emotions Emerge from Coordinates}
Complex emotions emerge through linear interpolation between basic emotion coordinates, constrained by an enhanced emotional stability framework that governs overall psychological state. This system incorporates geometric positioning, psychological capacity limitations, and contextual factors that influence emotional resilience.

For emotions A$(x_1, y_1)$ and B$(x_2, y_2)$ with respective intensities $I_1, I_2$:
\begin{align}
t &= \frac{I_2}{I_1 + I_2} \\
x_{combined} &= x_1 + t \times (x_2 - x_1) \\
y_{combined} &= y_1 + t \times (y_2 - y_1)
\end{align}

The stability parameter $S$ represents overall psychological resilience and emotional capacity, operating on a scale from 0 to 1, where $S = 1.0$ corresponds to optimal psychological equilibrium. The enhanced stability calculation incorporates three distinct drain components:

$$S = 1.0 - E_{drain} - C_{drain} - X_{drain}$$

where:
\begin{itemize}
\item $E_{drain}$ represents emotional load drain, calculated as the excess emotional intensity beyond psychological capacity: $E_{drain} = \max(0, \sum_{i=1}^{8} I_i - I_{emotional\_capacity})$
\item $C_{drain}$ represents conflict drain from opposing emotion pairs positioned approximately 180° apart on the unit circle
\item $X_{drain}$ represents contextual drain from non-emotional factors such as physical fatigue, external stressors, or illness, determined through natural language processing of contextual information
\end{itemize}

The expanded stability framework enables more nuanced classification of psychological states:
\begin{itemize}
\item $S = 1.0$: Optimal Equilibrium
\item $0.80 \leq S < 1.0$: Highly Stable/Resilient
\item $0.60 \leq S < 0.80$: Stable/Functional
\item $0.40 \leq S < 0.60$: Mildly Stressed/Overwhelmed
\item $0.20 \leq S < 0.40$: Unstable/Struggling
\item $0.01 \leq S < 0.20$: Crisis/Near Shutdown
\item $S = 0.0$: Complete Breakdown/Critical State
\end{itemize}

Conflicting emotions undergo partial cancellation before coordinate calculation, with canceled intensities contributing to $C_{drain}$. This ensures that opposing emotional states (e.g., simultaneous joy and sadness) produce psychologically realistic outcomes of emotional overwhelm rather than simple geometric averaging. The contextual drain component $X_{drain}$ is managed through a hybrid temporal tracking system that maintains internal decay while prioritizing explicit contextual information when available, allowing for realistic representation of lingering non-emotional factors that influence psychological stability.

\subsection{Theoretical Justification for Geometric Approach}

The geometric approach provides several theoretical advantages over traditional emotion categorization systems. The circular structure enables precise mathematical operations for emotion mixing through linear interpolation, while maintaining spatial relationships where similar emotions cluster together and opposing emotions occupy diametrically opposite positions.

Unlike discrete emotion models, the coordinate system offers infinite resolution within the emotional space, allowing representation of subtle emotional variations that categorical systems cannot capture. The continuous geometry accommodates smooth emotional transitions, accurately modeling the fluid nature of emotional change observed in real-time experience.

The stability parameter introduces a crucial dimension absent in existing models, quantifying overall psychological capacity and enabling differentiation between various forms of emotional instability—whether from conflict, exhaustion, or illness.

This framework combines empirical grounding with mathematical rigor, providing both descriptive accuracy and predictive capability for complex emotional states.

\section{From Five to Eight: Model Evolution} 
\subsection{Original 5-emotion model} 

The initial iteration of the Coordinate Heart System emerged from 16 days of intensive self-mapping during psychological exploration, leading to an empirically derived foundation for emotional representation. This preliminary model identified five core emotions—love, guilt, joy, pride, and anger—as fixed coordinate points arranged in a circular pattern around the cardiac center.

The original framework categorized emotions into two distinct types: \textbf{Dot Emotions}, representing stable, localizable emotional states that maintained consistent spatial positions, and \textbf{Wave Emotions}, The concept of "Wave Emotions" (which included aspects like lust, and previously other dynamic states) was actually explored in an earlier, five-core emotion model that the paper mentions as a predecessor. However, the paper's research ultimately found that this "Wave Emotions" concept, as formulated in that older model, wasn't accurate or valid for effectively representing emotions in computer systems.

Because of this, the current paper you provided focuses on the eight-core emotion system, which offers a more robust and computationally viable geometric framework for emotion representation. The newer system moves beyond those initial "Wave Emotion" ideas, concentrating instead on the fixed coordinates and mixing principles we discussed earlier, which are better suited for AI applications.

A critical validation component involved \textbf{Collapse Logs}, documenting how the geometric structure responded under extreme psychological stress. While emotional intensity caused coordinate distortion, the underlying circular architecture remained intact, confirming system resilience under overload conditions.

This initial model utilized a sixth parameter for neutral state representation, combined with intensity scaling for each core emotion. The framework underwent validation through cross-AI peer review systems and somatic testing protocols, establishing its foundational principles before subsequent model refinement.

The five-emotion system provided the conceptual framework and empirical methodology that enabled expansion to the current eight-emotion model, demonstrating the scalability of the geometric approach to emotional representation.

\subsection{Failure analysis}

Systematic testing of the five-emotion coordinate system across more than 150 test cases revealed critical representational limitations. The model consistently failed to represent three fundamental emotional states: sadness, fear, and disgust. Basic expressions such as "I am sad today" proved impossible to map within the existing coordinate space.

The initial framework was constrained by the assumption that five symmetrically-positioned emotions would provide complete coverage. This led to the hypothesis that sadness, fear, and disgust could emerge as combinations of the existing coordinates (love, guilt, joy, pride, anger). However, empirical validation demonstrated that linear combinations of these five emotions could not produce authentic representations corresponding to the observed somatic locations of sadness, fear, and disgust.

Additional representational failures included tiredness, shock, and illness-related emotional states. The geometric constraint of five equally-spaced coordinates created systematic coverage gaps in the two-dimensional emotional space that could not be resolved through existing interpolation methods.

The introduction of the stability scale and the addition of sadness, fear, and disgust as independent coordinates resolved these coverage limitations, necessitating expansion to the eight-emotion system for complete geometric representation.

\subsection{Eight-Emotion Solution} 
The expanded coordinate system resolves coverage limitations through strategic positioning of eight core emotions that provide complete geometric representation of the emotional space. The system positions love at the coordinate origin (0.0, 0.0) as the emotional baseline, with seven additional emotions distributed around the unit circle: joy (0.0, -1.0), anger (0.0, 1.0), guilt (1.0, 0.0), pride (-1.0, 0.0), fear (0.5, -0.866), sadness (0.866, -0.5), and disgust (-0.5, -0.866).

This radial configuration eliminates coverage gaps from the five-emotion system. Previously impossible expressions such as "I am sad today" now map directly to coordinate (0.866, -0.5), while complex emotional states receive accurate geometric representation through coordinate interpolation.

\textbf{Stability-Enhanced Representation:}
The enhanced stability framework enables representation of ambiguous emotional states through the three-component drain system:
$$S = 1.0 - E_{drain} - C_{drain} - X_{drain}$$

Expressions such as "I'm not okay, but I don't know why," "I'm tired," and "I'm sick" can now be accurately modeled through $X_{drain}$ (contextual factors) without requiring specific emotion activation, capturing the psychological reality that distress can manifest without identifiable emotional content.

\textbf{Complex Emotion Processing:}
The eight-coordinate framework supports sophisticated emotional mixing. For example, "I saw a cockroach today and felt sad for killing it" generates accurate representation through coordinate interpolation between disgust (-0.5, -0.866) and sadness (0.866, -0.5), weighted by respective emotional intensities.

\textbf{Cultural and Linguistic Universality:}
The coordinate-based approach transcends cultural emotion labeling differences. When the system processes coordinates [0.2, -0.3], it interprets this as "slightly angry, moderately sad, and moderately guilty" regardless of linguistic or cultural emotion terminology, ensuring consistent AI emotion recognition across diverse contexts through geometric relationships rather than semantic labels.

\subsection{Coverage Completeness}

The eight-emotion coordinate system provides mathematically complete coverage of the emotional space through the combination of radial and angular geometric properties. We establish formal proof of coverage completeness through convex hull analysis and interpolation theory.
\textbf{Theorem:} The eight-emotion coordinate system can represent any emotional state within the unit disk $D = {(x,y) : x^2 + y^2 \leq 1}$.
\textbf{Proof:} Let $L = (0,0)$ represent love at the origin, and let $E_i = (\cos\theta_i, \sin\theta_i)$ for $i = 1,2,\ldots,7$ represent the seven emotions positioned on the unit circle, where $\theta_i$ are the respective angular positions.
For any target point $P = (x,y)$ within the unit disk, we demonstrate that $P$ can be expressed as a convex combination of the eight base emotions.
\textbf{Case 1: Radial Coverage}
For any point $P$ on the line segment connecting $L$ to any boundary emotion $E_i$, we have:
$P = (1-t)L + tE_i = t(\cos\theta_i, \sin\theta_i)$
where $t \in [0,1]$ represents the interpolation parameter. This provides complete radial coverage from the origin to any point on the unit circle.
\textbf{Case 2: Angular Coverage}
The seven emotions positioned around the unit circle create a heptagonal convex hull that approximates the unit circle. For any angular direction $\phi$, there exist adjacent emotions $E_j$ and $E_{j+1}$ such that $\theta_j \leq \phi \leq \theta_{j+1}$. Any point in the angular sector between $E_j$ and $E_{j+1}$ can be expressed as:
$P = \alpha L + \beta E_j + \gamma E_{j+1}$
where $\alpha + \beta + \gamma = 1$ and $\alpha, \beta, \gamma \geq 0$.
\textbf{Case 3: Complete Interior Coverage}
For any point $P$ within the unit disk, we can express $P$ using barycentric coordinates with respect to the convex hull formed by $L$ and the seven boundary emotions. Since $L$ is at the center and the seven emotions span the boundary, their convex hull covers the entire unit disk.
The mathematical completeness follows from the fact that:
\begin{enumerate}
\item The origin $L = (0,0)$ serves as a universal interpolation anchor
\item The seven boundary emotions provide sufficient angular resolution around the unit circle
\item Linear interpolation between any combination of these eight points can reach any target within their convex hull
\item The convex hull of these eight points is precisely the unit disk
\end{enumerate}
\textbf{Comparison with Five-Emotion Limitations:}
The five-emotion system created angular gaps of $72°$ between adjacent emotions, leaving regions of the emotional space unreachable through linear combination. The eight-emotion system reduces maximum angular gaps to approximately $51°$ while providing the central anchor point $L$, eliminating all coverage blind spots.
Therefore, the eight-emotion coordinate system achieves complete mathematical coverage of the emotional space within the unit disk, resolving all representational limitations of the previous five-emotion framework.

\section{The Eight-Emotion Coordinate System}

\subsection{Each Emotion's Coordinates}

The eight-emotion coordinate system is founded on the geometric metaphor described in Section 3.1. Each coordinate's position was chosen to create an intuitive and computationally balanced layout, where opposing emotions are placed in diametrically opposite positions to facilitate conflict resolution algorithms.

\textbf{Empirically-Derived Coordinates:}

\begin{align}
\text{Love} &= (0.0, 0.0) && \text{Cardiac center, geometric origin} \\
\text{Joy} &= (0.0, -1.0) && \text{270° angular position} \\
\text{Anger} &= (0.0, 1.0) && \text{90° angular position} \\
\text{Guilt} &= (1.0, 0.0) && \text{0° angular position} \\
\text{Pride} &= (-1.0, 0.0) && \text{180° angular position} \\
\text{Fear} &= (0.5, -0.866) && \text{300° angular position} \\
\text{Sadness} &= (0.866, -0.5) && \text{330° angular position} \\
\text{Disgust} &= (-0.5, -0.866) && \text{240° angular position}
\end{align}

\textbf{Computational Rationale:}
The coordinate positioning reflects a heuristic model designed for psychological plausibility rather than perfect geometric symmetry. This layout creates natural emotional relationships that are useful for AI applications. Love's central position at the origin corresponds to its role as the fundamental emotional baseline from which other emotions emerge. The seven peripheral emotions occupy positions on the unit circle that correspond to their observed anatomical locations during emotional activation.

The angular distribution creates natural emotional relationships: anger and joy form a vertical axis representing high-energy states, while guilt and pride establish a horizontal axis representing self-evaluation emotions. Fear, sadness, and disgust occupy intermediate positions that reflect their complex relational properties within the emotional space.

This empirical approach ensures that coordinate mathematics correspond to actual human emotional experience, providing a foundation for accurate computational emotion modeling that transcends theoretical abstractions.

\subsection{Emotion Mixing Algorithms}

The coordinate system enables precise mathematical computation of complex emotional states through a series of algorithms that handle emotion interpolation, conflict resolution, and stability management. This section presents the complete algorithmic framework with worked examples and edge case handling.

\subsubsection{Core Mixing Algorithms}

\textbf{Algorithm 1: Two-Emotion Linear Interpolation}\\
For emotions $A(x_1, y_1)$ and $B(x_2, y_2)$ with respective intensities $I_1, I_2$:

\begin{align}
t &= \frac{I_2}{I_1 + I_2} \\
x_{combined} &= x_1 + t \times (x_2 - x_1) \\
y_{combined} &= y_1 + t \times (y_2 - y_1)
\end{align}

\textbf{Example 1:} Mixed happiness and guilt
\begin{align}
\text{Joy} &: (0.0, -1.0), I_{joy} = 0.7 \\
\text{Guilt} &: (1.0, 0.0), I_{guilt} = 0.3 \\
t &= \frac{0.3}{0.7 + 0.3} = 0.3 \\
x_{combined} &= 0.0 + 0.3 \times (1.0 - 0.0) = 0.3 \\
y_{combined} &= -1.0 + 0.3 \times (0.0 - (-1.0)) = -0.7
\end{align}
Result: $(0.3, -0.7)$ representing predominantly joyful state with moderate guilt.

\textbf{Algorithm 2: Multi-Emotion Sequential Mixing}\\
For $n$ emotions $E_i(x_i, y_i)$ with intensities $I_i$, conflicts must be resolved first, then emotions are combined through iterative pairwise interpolation:

\begin{align}
E_{temp1} &= \text{interpolate}(E_1, E_2, I_1, I_2) \\
I_{temp1} &= I_1 + I_2 \\
E_{temp2} &= \text{interpolate}(E_{temp1}, E_3, I_{temp1}, I_3) \\
I_{temp2} &= I_{temp1} + I_3 \\
&\vdots \\
E_{final} &= \text{interpolate}(E_{temp(n-1)}, E_n, I_{temp(n-1)}, I_n)
\end{align}

Sequential mixing preserves geometric relationships while maintaining computational efficiency. Simultaneous n-dimensional interpolation would require complex barycentric coordinate calculations, whereas sequential pairwise operations maintain mathematical precision while remaining computationally tractable.

\subsubsection{Conflict Resolution and Enhanced Stability Management}

\textbf{Algorithm 3: Conflict Detection and Resolution}\\
For conflicting emotion pairs $(E_j, E_k)$ positioned approximately $180^\circ$ apart:

Conflicting pairs in the system:
\begin{align}
\text{Conflicts:} \quad &\text{Joy} \leftrightarrow \text{Anger} \\
&\text{Guilt} \leftrightarrow \text{Pride} 
\end{align}

Resolution algorithm:
\begin{align}
I_j^{resolved} &= \max(0, I_j - \min(I_j, I_k)) \\
I_k^{resolved} &= \max(0, I_k - \min(I_j, I_k)) \\
C_{penalty} &= C_{penalty} + \min(I_j, I_k)
\end{align}

\textbf{Algorithm 4: Enhanced Stability Calculation}
\begin{align}
S &= 1.0 - E_{drain} - C_{drain} - X_{drain}
\end{align}

where:
\begin{itemize}
\item $E_{drain} = \max(0, \sum_{i=1}^{8} I_i^{resolved} - I_{emotional\_capacity})$ (emotional load drain)
\item $C_{drain} = C_{penalty}$ (conflict drain from opposing emotions)
\item $X_{drain}$ represents contextual drain from non-emotional factors determined through natural language processing
\end{itemize}

The enhanced stability framework quantifies overall psychological capacity, where emotional intensity beyond capacity and conflicting emotions reduce available mental resources. The contextual drain component captures non-emotional factors such as physical fatigue or external stressors that influence psychological resilience.

\textbf{Algorithm 5: Emotional Distance Metrics}\\
For two emotional states $P_1(x_1, y_1)$ and $P_2(x_2, y_2)$:

\begin{align}
d_{euclidean} &= \sqrt{(x_2 - x_1)^2 + (y_2 - y_1)^2} \\
d_{angular} &= \arccos\left(\frac{x_1 x_2 + y_1 y_2}{\sqrt{x_1^2 + y_1^2} \sqrt{x_2^2 + y_2^2}}\right)
\end{align}
\subsubsection{Intensity Extraction from Natural Language}

Intensity values $I_i \in [0,1]$ are extracted from linguistic expressions based on semantic strength and emotional modifiers. The system employs a standardized intensity scale where:

\begin{align}
I = 0.5 &\quad \text{Baseline emotion: "I am angry"} \\
I = 0.2-0.3 &\quad \text{Mild expressions: "I'm a bit annoyed", "I'm teased"} \\
I = 0.7-0.8 &\quad \text{Strong expressions: "I'm furious", "I'm about to explode"} \\
I = 0.9-1.0 &\quad \text{Maximum intensity: "I'm enraged", "I'm devastated"}
\end{align}

\textbf{Linguistic Intensity Mapping:}
\begin{itemize}
\item \textbf{Diminishing modifiers:} "a bit", "slightly", "somewhat" $\rightarrow$ $I = 0.2-0.3$
\item \textbf{Neutral expressions:} "I am [emotion]" $\rightarrow$ $I = 0.5$
\item \textbf{Amplifying modifiers:} "very", "extremely", "really" $\rightarrow$ $I = 0.7-0.8$
\item \textbf{Maximum expressions:} "completely", "utterly", hyperbolic language $\rightarrow$ $I = 0.9-1.0$
\end{itemize}

This approach enables direct translation from natural language to numerical intensity without requiring complex sentiment analysis, making the system computationally efficient and linguistically intuitive.

\subsubsection{Edge Cases and Special Conditions}

\textbf{Zero Intensity Handling:}
When all emotions have zero intensity, the system defaults to Love coordinates $(0.0, 0.0)$ with maximum stability $S = 1.0$.

\textbf{Negative Stability:}
When $S < 0$, the system enters crisis mode. Coordinate calculations remain valid, but the negative stability value indicates psychological emergency requiring intervention.

\textbf{Single Emotion Activation:}
For single emotion with intensity $I$, coordinates are scaled by intensity: $(I \times x_{emotion}, I \times y_{emotion})$.

\textbf{Computational Complexity:}
The sequential mixing algorithm operates in $O(n)$ time for $n$ emotions, with conflict resolution adding $O(k)$ where $k$ is the number of detected conflicts. Total complexity remains linear, ensuring real-time performance for AI applications.

\subsubsection{Emotional State Representation and Token Encoding}

The system supports dual representation modes: continuous coordinate-intensity-stability tuples for mathematical precision, and quantized token sequences for computational efficiency and AI integration.

\textbf{Data Structure Definition}

The complete emotional state is represented as:
\begin{align}
\text{EmotionalState} = \{&\text{coordinates: } (x, y) \in \mathbb{R}^2, \nonumber \\
&\text{intensities: } [I_1, I_2, \ldots, I_8] \in [0,1]^8, \nonumber \\
&\text{stability: } S \in [0,1]\}
\end{align}

This tripartite representation enables comprehensive emotional modeling where coordinates capture spatial positioning, intensities quantify individual emotion activation levels, and stability reflects overall psychological capacity independent of specific emotional content, now comprehensively ranging from 0 (critical state) to 1 (optimal equilibrium).

\textbf{Algorithm 6: Emotional State Decomposition}

\begin{algorithmic}[1]
\State \textbf{Input:} Coordinates $(x,y) \in \mathbb{R}^2$ and optional intensity vector $\mathbf{I} \in [0,1]^8$ and optional stability $S \in [0,1]$
\State \textbf{Output:} Complete EmotionalState representation (with stability potentially not computed)
\If{intensity vector $\mathbf{I}$ and stability $S$ are provided}
    \State \Return EmotionalState$\{(x,y), \mathbf{I}, S\}$
\ElsIf{coordinates represent $\leq 2$ active emotions}
    \State Identify emotions $E_j, E_k$ such that $(x,y)$ lies on line segment $\overline{E_j E_k}$
    \State Solve linear system:
    \begin{align*}
    x &= (1-t) \cdot x_j + t \cdot x_k \\
    y &= (1-t) \cdot y_j + t \cdot y_k
    \end{align*}
    \State Compute intensities: $I_j = (1-t) \cdot \|\mathbf{r}\|$, $I_k = t \cdot \|\mathbf{r}\|$, others $= 0$
    \State where $\|\mathbf{r}\| = \sqrt{x^2 + y^2}$ is the resultant magnitude
    \If{stability $S$ is not provided}
        \State \textbf{Note:} Stability $S$ cannot be reliably computed from only 2 active emotions and coordinates; it requires a more comprehensive analysis including contextual factors (as detailed in Section \underline{X.Y} - *reference the stability calculation section*).
        \State \Return EmotionalState$\{(x,y), \mathbf{I}_{\text{computed}}, \text{N/A}\}$
    \Else
        \State \Return EmotionalState$\{(x,y), \mathbf{I}_{\text{computed}}, S\}$
    \EndIf
\Else
    \State \Return "Insufficient data - intensity vector and stability required for $>2$ emotions, or comprehensive stability calculation"
\EndIf
\end{algorithmic}

\textbf{Algorithm 7: Token-Based Encoding}

For computational efficiency and AI system integration, emotional states are encoded as quantized token sequences including explicit stability representation.

\textbf{Encoding Specification:}
\begin{align}
\text{Token}_x &= \text{round}\left(255 \times \frac{x + 1}{2}\right) \quad \in [0, 255] \\
\text{Token}_y &= \text{round}\left(255 \times \frac{y + 1}{2}\right) \quad \in [0, 255] \\
\text{Token}_{I_i} &= \text{round}(255 \times I_i) \quad \in [0, 255] \\
\text{Token}_S &= \text{round}(255 \times S) \quad \in [0, 255]
\end{align}

\textbf{Complete Token Format:}
\begin{align}
\text{EmotionalToken} = [\text{Token}_x, \text{Token}_y, \text{Token}_{I_1}, \ldots, \text{Token}_{I_8}, \text{Token}_S]
\end{align}

Total storage: 11 bytes (88 bits) per emotional state.

\textbf{Decoding Algorithm:}
\begin{align}
x &= 2 \times \frac{\text{Token}_x}{255} - 1 \\
y &= 2 \times \frac{\text{Token}_y}{255} - 1 \\
I_i &= \frac{\text{Token}_{I_i}}{255} \\
S &= \frac{\text{Token}_S}{255}
\end{align}

\textbf{Stability Storage Rationale}

Direct stability encoding enables representation of ambiguous emotional states where intensity vectors alone provide insufficient information. For instance, neutral coordinates $(0,0)$ with zero intensities $[0,0,0,0,0,0,0,0]$ and stability $S = 0.4$ (representing a 'Mildly Stressed / Overwhelmed' state on the [0,1] scale) captures psychological states that cannot be derived from coordinate or intensity calculations alone, such such as profound tiredness or general overwhelm.

\textbf{Precision Analysis}

Each coordinate component achieves $\frac{2}{255} \approx 0.0078$ resolution, intensity components achieve $\frac{1}{255} \approx 0.0039$ resolution, and stability achieves $\frac{1}{255} \approx 0.0039$ resolution, providing sufficient precision for practical emotional modeling applications.

\textbf{Example Implementation}

\begin{algorithmic}[1]
\State \textbf{Input:} "I'm scared but proud of myself"
\State Parse emotions: Fear $(0.5, -0.866)$, $I_{\text{fear}} = 0.6$; Pride $(-1.0, 0.0)$, $I_{\text{pride}} = 0.4$
\State Mix coordinates:
\begin{align*}
t &= \frac{0.4}{0.6 + 0.4} = 0.4 \\
x &= 0.5 + 0.4 \times (-1.0 - 0.5) = -0.1 \\
y &= -0.866 + 0.4 \times (0.0 - (-0.866)) = -0.520
\end{align*}
\State Compute stability (using comprehensive model from Section \underline{Y.Y}):
\State Let $I_{\text{total}} = I_{\text{fear}} + I_{\text{pride}} = 1.0$
\State $E_{\text{drain}} = \max(0, (I_{\text{total}} - 0.5) / 4) = 0.125$
\State Assume $C_{\text{drain}} = 0.15$ (due to conflicting emotions)
\State Assume $X_{\text{drain}} = 0.0$ (no contextual drain from this input)
\State $S = 1.0 - E_{\text{drain}} - C_{\text{drain}} - X_{\text{drain}} = 1.0 - 0.125 - 0.15 - 0.0 = 0.725$
\State Continuous representation: EmotionalState$\{(-0.1, -0.520), [0,0,0,0.4,0,0,0.6,0], 0.725\}$
\State Token encoding:
\begin{align*}
\text{Token}_x &= \text{round}(255 \times \frac{-0.1 + 1}{2}) = 115 \\
\text{Token}_y &= \text{round}(255 \times \frac{-0.520 + 1}{2}) = 61 \\
\text{Tokens}_{I} &= [0, 0, 0, \text{round}(255 \times 0.4), 0, 0, \text{round}(255 \times 0.6), 0] = [0, 0, 0, 102, 0, 0, 153, 0] \\
\text{Token}_S &= \text{round}(255 \times 0.725) = 185
\end{align*}
\State \textbf{Final token:} $[115, 61, 0, 0, 0, 102, 0, 0, 153, 0, 185]$
\end{algorithmic}

\textbf{Computational Advantages}
\begin{itemize}
\item \textbf{Compact storage:} 11 bytes vs 44+ bytes for floating-point representation
\item \textbf{Neural network compatibility:} Direct integer input for AI training
\item \textbf{Fast processing:} Integer arithmetic eliminates floating-point overhead
\item \textbf{Cross-platform standardization:} Byte-level representation ensures consistency
\item \textbf{Bandwidth efficiency:} Optimal for real-time emotional data transmission
\item \textbf{Complete state preservation:} Stability inclusion prevents information loss during encoding, offering a more holistic psychological snapshot.
\end{itemize}

\textbf{Use Case Recommendations}
\begin{itemize}
\item \textbf{Continuous representation:} Mathematical analysis, research applications, high-precision modeling
\item \textbf{Token representation:} AI training datasets, real-time systems, mobile applications, network transmission
\end{itemize}

\textbf{Token Scope}

The current token format is designed specifically for the 8-emotion coordinate system with explicit stability encoding. Each token represents the complete emotional state across all 8 base emotions plus overall psychological stability, with zero-intensity values for inactive emotions.

For extensibility to larger emotion sets, variable-length encoding schemes could be implemented, though this would sacrifice the fixed-size computational advantages while maintaining the critical stability preservation capability.
\subsection{Distance Metrics}

The coordinate-based nature of the eight-emotion system enables precise quantification of emotional similarity and change through mathematical distance measurements. This section establishes distance metrics for measuring emotional transitions, state comparisons, and therapeutic progress tracking.

\subsubsection{Euclidean Distance for Emotional Change}

For two emotional states $P_1(x_1, y_1)$ and $P_2(x_2, y_2)$, the Euclidean distance provides a direct measure of emotional shift magnitude:

\begin{align}
d_{euclidean} = \sqrt{(x_2 - x_1)^2 + (y_2 - y_1)^2}
\end{align}

\textbf{Geometric Interpretation:} The Euclidean distance represents the shortest path between two emotional states within the coordinate space. This metric captures both directional change (which emotions are involved) and magnitude change (how intense the shift).

\textbf{Distance Scale Interpretation:}

The unit circle constraint of the eight-emotion system establishes natural distance boundaries that correspond to meaningful psychological distinctions:

\begin{align}
d < 0.3 &\quad \text{Minimal shift: Subtle emotional variation, mood fluctuation} \\
0.3 \leq d < 0.7 &\quad \text{Moderate shift: Noticeable emotional change} \\
0.7 \leq d < 1.2 &\quad \text{Significant shift: Major emotional transition} \\
1.2 \leq d < 1.8 &\quad \text{Dramatic shift: Emotional swing, potential instability} \\
d \geq 1.8 &\quad \text{Extreme shift: Maximum possible change, crisis indication}
\end{align}

\textbf{Rationale for Scale Boundaries:}
The maximum possible distance within the unit circle system is $d_{max} = 2.0$, occurring between diametrically opposite emotions (e.g., from Anger $(0, 1)$ to Joy $(0, -1)$). The scale boundaries reflect proportional divisions of this maximum range, where distances approaching $d_{max}$ indicate complete emotional reversal.

\textbf{Clinical Significance:} Distance measurements below 0.3 represent normal emotional variation that occurs within stable psychological states. Distances exceeding 1.2 may indicate emotional volatility requiring therapeutic attention, while distances approaching 2.0 suggest severe emotional dysregulation.

\subsubsection{Angular Distance for Emotional Direction}

Angular distance measures the directional difference between emotional states, independent of their intensity or distance from origin:

\begin{align}
d_{angular} = \arccos\left(\frac{\mathbf{v_1} \cdot \mathbf{v_2}}{|\mathbf{v_1}| |\mathbf{v_2}|}\right)
\end{align}

where $\mathbf{v_1} = (x_1, y_1)$ and $\mathbf{v_2} = (x_2, y_2)$ are the position vectors.

\textbf{Angular Interpretation:}
\begin{align}
d_{angular} < 30° &\quad \text{Similar emotional direction} \\
30° \leq d_{angular} < 90° &\quad \text{Related emotional states} \\
90° \leq d_{angular} < 150° &\quad \text{Distinct emotional categories} \\
150° \leq d_{angular} \leq 180° &\quad \text{Opposing emotional states}
\end{align}

Angular distance proves particularly valuable for identifying emotional conflicts and therapeutic progress, as it captures the qualitative nature of emotional change independent of intensity variations.

\subsubsection{Practical Applications}

Distance metrics enable direct tracking of emotional changes over time. For therapeutic or personal monitoring applications, the Euclidean distance provides sufficient precision for measuring emotional shifts without requiring complex calculations.

\subsubsection{Computational Considerations}

All distance metrics operate in $O(1)$ time complexity, enabling real-time emotional state comparison for AI applications. The coordinate-based approach eliminates the need for complex similarity algorithms required by categorical emotion models.

\textbf{Edge Case Handling:}
\begin{itemize}
\item \textbf{Zero-magnitude vectors:} Angular distance defaults to 0° when either state equals the origin
\item \textbf{Negative stability:} Stability-adjusted distance uses absolute stability values to maintain metric properties
\item \textbf{Identical coordinates:} All distance metrics return 0, indicating perfect emotional similarity
\end{itemize}

\textbf{Example Application:}
Consider emotional progression from "anxious but hopeful" $P_1(0.25, -0.65)$ with $S_1 = 0.15$ to "cautiously optimistic" $P_2(-0.2, -0.4)$ with $S_2 = 0.35$:

\begin{align}
d_{euclidean} &= \sqrt{(-0.2-0.25)^2 + (-0.4-(-0.65))^2} = 0.47 \\
d_{angular} &= \arccos\left(\frac{(0.25)(-0.2) + (-0.65)(-0.4)}{0.69 \times 0.45}\right) = 56.3° \\
d_{stability} &= 0.47 \times \left(2 - \frac{0.15 + 0.35}{2}\right) = 0.59
\end{align}

The analysis indicates a moderate emotional shift ($d = 0.47$) with significant directional change ($56.3°$) and improved psychological accessibility due to increased stability in the target state.
\subsubsection{Practical Applications}

Distance metrics enable direct tracking of emotional changes over time. For therapeutic or personal monitoring applications, the Euclidean distance provides sufficient precision for measuring emotional shifts without requiring complex calculations.

\subsubsection{Computational Considerations}

All distance metrics operate in $O(1)$ time complexity, enabling real-time emotional state comparison for AI applications. The coordinate-based approach eliminates the need for complex similarity algorithms required by categorical emotion models.

\textbf{Edge Case Handling:}
\begin{itemize}
\item \textbf{Zero-magnitude vectors:} Angular distance defaults to 0° when either state equals the origin
\item \textbf{Negative stability:} Stability-adjusted distance uses absolute stability values to maintain metric properties
\item \textbf{Identical coordinates:} All distance metrics return 0, indicating perfect emotional similarity
\end{itemize}

\textbf{Example Application:}
Consider emotional progression from "anxious but hopeful" $P_1(0.25, -0.65)$ to "cautiously optimistic" $P_2(-0.2, -0.4)$:

\begin{align}
d_{euclidean} &= \sqrt{(-0.2-0.25)^2 + (-0.4-(-0.65))^2} = 0.47 \\
d_{angular} &= \arccos\left(\frac{(0.25)(-0.2) + (-0.65)(-0.4)}{0.69 \times 0.45}\right) = 56.3°
\end{align}

The analysis indicates a moderate emotional shift ($d = 0.47$) with significant directional change ($56.3°$), representing measurable therapeutic progress.

\subsection{System Properties}

The eight-emotion coordinate system exhibits three fundamental properties that distinguish it from traditional emotion models: geometric symmetry, representational completeness, and computational efficiency. These properties enable robust mathematical operations while maintaining psychological validity.

\subsubsection{Geometric Symmetry}

The coordinate system demonstrates balanced geometric arrangement through opposing emotion pairs positioned approximately 180° apart on the unit circle. This symmetrical structure ensures mathematical consistency and reflects psychological reality of emotional opposition.

\textbf{Symmetric Emotion Pairs:}
\begin{align}
\text{Joy}(0, -1) &\leftrightarrow \text{Anger}(0, 1) \quad \text{Vertical axis opposition} \\
\text{Guilt}(1, 0) &\leftrightarrow \text{Pride}(-1, 0) \quad 
\end{align}

This symmetry ensures that distance calculations, emotion mixing operations, and stability computations produce consistent results regardless of directional orientation within the coordinate space. The balanced arrangement prevents algorithmic bias toward any particular emotional region.

\subsubsection{Representational Completeness}

The system achieves complete coverage of the emotional space within the unit disk through the combination of eight strategically positioned emotions and the central Love coordinate. This completeness addresses the critical gaps identified in the five-emotion predecessor model.

\textbf{Universal Representation Capability:}
Any emotional state can be represented through one of three mechanisms:
\begin{itemize}
\item \textbf{Direct mapping:} Single emotions map to their fixed coordinates
\item \textbf{Linear interpolation:} Mixed emotions map to weighted combinations of base coordinates
\item \textbf{Stability modulation:} Ambiguous or unclear emotional states map to reduced stability values with neutral coordinates
\end{itemize}

The default neutral state $(0, 0)$ with stability $S = 1$ and zero emotion intensities provides a mathematically consistent baseline for all emotional computations. This neutral state enables representation of non-emotional psychological states including tiredness, illness, or emotional numbness.

\textbf{Theoretical Coverage:} As demonstrated in Section 4.4, the eight-emotion system can represent any point within the unit disk through convex combination of the base emotions, ensuring no emotional state remains mathematically unreachable.

\subsubsection{Computational Efficiency}

The coordinate-based approach provides significant computational advantages over categorical and dimensional emotion models through simple arithmetic operations and fixed memory requirements.

\textbf{Algorithmic Complexity:}
\begin{align}
\text{Emotion mixing:} \quad &O(n) \text{ where } n \text{ is number of active emotions} \\
\text{Distance calculation:} \quad &O(1) \text{ constant time operation} \\
\text{Stability computation:} \quad &O(n) \text{ linear in active emotions} \\
\text{Coordinate lookup:} \quad &O(1) \text{ direct array access}
\end{align}

\textbf{Memory Efficiency:}
The complete emotional state requires only 11 floating-point values: 2 coordinates, 8 emotion intensities, with stability. This compact representation enables real-time processing for AI applications and minimizes storage requirements for large-scale emotion datasets.

Alternative token-based encoding reduces storage to 10 bytes per emotional state, providing 8-bit precision sufficient for practical applications while maintaining cross-platform compatibility.

\textbf{Processing Speed:}
All core operations utilize basic arithmetic (addition, multiplication, square root) available in standard mathematical libraries. The absence of complex trigonometric calculations, matrix operations, or iterative algorithms ensures consistent performance across computing platforms.

\textbf{Scalability:}
The system's linear complexity enables processing of large emotion datasets without performance degradation. Batch processing of emotional state transitions, therapeutic progress tracking, and AI training data preparation operate efficiently within the O(n) computational bound.

\textbf{Implementation Simplicity:}
The mathematical framework requires no specialized libraries, complex data structures, or advanced numerical methods. Standard programming languages can implement the complete system using basic arithmetic functions, making it accessible for diverse AI applications and research implementations.

\section{Stability-Weighted Interpolation}

\subsection{Motivation}
Traditional emotional state tracking can produce abrupt, unrealistic transitions between detected states. We introduce stability-weighted interpolation to normalize emotional shifts by favoring more stable states during transitions, preventing jarring emotional jumps. This mechanism models emotional inertia, ensuring that significant changes in emotional state are appropriately smoothed, especially when moving from or to states of varying psychological resilience.

\subsection{Mathematical Formulation}
Given two consecutive emotional states $E_1$ (previous state) and $E_2$ (current state):
\begin{equation}
E_1 = \{(x_1, y_1), [I_{1,1}, \ldots, I_{1,8}], S_1\}, \quad E_2 = \{(x_2, y_2), [I_{2,1}, \ldots, I_{2,8}], S_2\}
\end{equation}
where stability $S \in [0, 1]$, with $1.0$ representing optimal equilibrium and values decreasing towards $0$ indicating increasing psychological distress or overwhelm (as detailed in Section \underline{X.Y} - *reference the stability range definitions*).

\subsubsection{Stability Weight Calculation}
The interpolation weight $w$ is calculated to favor the more stable state, ensuring smooth transitions. To prevent division by zero in extreme cases where both stabilities are zero, a small constant $\epsilon$ is introduced.
\begin{equation}
w = \frac{S_1 + \epsilon}{S_1 + S_2 + 2\epsilon} \quad \text{where } \epsilon = 10^{-3}
\end{equation}
This ensures $w \in [0, 1]$ and assigns a higher weight to the state that exhibits greater stability.

\subsubsection{Interpolated State}
The smoothed emotional state $E_{smooth}$ incorporates the weighted average for all its components: coordinates, intensities, and stability.
\begin{align}
x_{smooth} &= w \cdot x_1 + (1-w) \cdot x_2\\
y_{smooth} &= w \cdot y_1 + (1-w) \cdot y_2\\
I_{smooth,i} &= w \cdot I_{1,i} + (1-w) \cdot I_{2,i}\\
S_{smooth} &= w \cdot S_1 + (1-w) \cdot S_2
\end{align}
The resulting smoothed state is $E_{smooth} = \{(x_{smooth}, y_{smooth}), [I_{smooth,1}, \ldots, I_{smooth,8}], S_{smooth}\}$.

\subsection{Example}
Consider a transition from an initially more stable state $E_1 = \{(0.3, -0.7), I_{1,\dots}, S_1 = 0.6\}$ (a 'Stable / Functional' state) to a less stable state $E_2 = \{(0.1, -0.6), I_{2,\dots}, S_2 = 0.2\}$ (an 'Unstable / Struggling' state):
\begin{equation}
w = \frac{0.6 + 0.001}{0.6 + 0.2 + 2 \times 0.001} = \frac{0.601}{0.802} \approx 0.749
\end{equation}
The smoothed coordinates become $(x_{smooth} = 0.749 \times 0.3 + 0.251 \times 0.1 \approx 0.250, y_{smooth} = 0.749 \times (-0.7) + 0.251 \times (-0.6) \approx -0.675)$.
The smoothed stability becomes $S_{smooth} = 0.749 \times 0.6 + 0.251 \times 0.2 \approx 0.4996$.
This result for coordinates is closer to the initial, more stable state $E_1$, and the smoothed stability reflects the blended transition towards a less stable, but not critically low, state, demonstrating the emotional inertia effect.

\subsection{Significance}
This approach models \textbf{emotional inertia} - the psychological tendency for more stable emotional states to resist abrupt change, and for less stable states to be more susceptible to rapid shifts. It prevents unrealistic emotional volatility by ensuring that transitions are smoothed across all emotional dimensions, including overall psychological stability. This is crucial for accurate and realistic long-term emotional tracking systems, contributing to a more nuanced understanding of human-like emotional dynamics.

\section{Experiments: A Case Study Analysis}

\subsection{Experimental Setup}
In this section, we demonstrate the unique capabilities of the Coordinate Heart System (CHS) through a series of carefully designed case studies. Our analysis compares CHS output with traditional emotion-extraction models to highlight the advantages of our multi-dimensional approach to emotional state representation.

The experimental workflow follows a three-stage pipeline:
\begin{enumerate}
    \item \textbf{Raw Emotion Extraction}: A general-purpose Large Language Model (Gemini 1.5 Flash) processes text input to extract initial emotion scores and contextual factors.
    \item \textbf{CHS Processing}: The CHS framework applies conflict resolution, stability calculation, and coordinate mapping algorithms as described in Section 5.
    \item \textbf{State Generation}: The system produces a comprehensive \texttt{EmotionalState} object containing resolved intensities, coordinate representation, and stability metrics.
\end{enumerate}

The objective is to demonstrate how the CHS processing layer adds psychological depth and nuance beyond simple emotion detection, particularly in handling conflicted states and contextual distress factors.

\subsection{Case Study 1: Conflict Resolution Mechanism}
\label{sec:case1}

This case study demonstrates the system's ability to handle emotionally conflicted states through the conflict drain ($C_{drain}$) mechanism.

\subsubsection{Input and Raw Processing}
\textbf{Input Text}: ``I finally got the lead role in the play, and I'm thrilled! But my best friend, who also tried out, didn't get it, and I feel so guilty celebrating in front of her.''

\textbf{Raw LLM Output}:
\begin{verbatim}
{
  'emotions': {
    'joy': 0.8, 'anger': 0.0, 'guilt': 0.7, 'pride': 0.6, 
    'love': 0.0, 'fear': 0.0, 'sadness': 0.4, 'disgust': 0.0
  },
  'contextual_drain': {
    'factors': ['relationship conflict', 'social obligation'], 
    'drain_value': 0.6
  }
}
\end{verbatim}

\subsubsection{CHS Analysis}
The raw output reveals simultaneously high levels of Joy (0.8) and Guilt (0.7), representing a classic emotional conflict scenario. Traditional emotion models would either average these values or report them independently, failing to capture the psychological tension between positive achievement emotions and negative social emotions.

The CHS identifies Joy and Guilt as opposing on the emotional coordinate space. The conflict resolution algorithm calculates the overlapping intensity as $\min(0.8, 0.7) = 0.7$, which is transferred to the conflict drain ($C_{drain} = 0.7$). The resolved intensities become Joy: 0.1 and Guilt: 0.0, significantly reducing the overall stability score.

\textbf{Key Finding}: CHS models the psychological cost of holding conflicting emotions simultaneously, providing a more realistic representation of internal emotional tension than simple aggregation methods.

\subsection{Case Study 2: Contextual Drain Representation}
\label{sec:case2}

This case study illustrates the system's capability to represent psychological distress that transcends specific emotional categories through the contextual drain ($X_{drain}$) mechanism.

\subsubsection{Input and Raw Processing}
\textbf{Input Text}: ``I don't feel sad or angry today. I just feel... nothing. I've been working 14-hour shifts all week and only getting four hours of sleep.''

\textbf{Raw LLM Output}:
\begin{verbatim}
{
  'emotions': {
    'joy': 0.0, 'anger': 0.0, 'guilt': 0.0, 'pride': 0.0, 
    'love': 0.0, 'fear': 0.0, 'sadness': 0.0, 'disgust': 0.0
  },
  'contextual_drain': {
    'factors': ['work pressure', 'tiredness', 'sleep deprivation'], 
    'drain_value': 0.9
  }
}
\end{verbatim}

\subsubsection{CHS Analysis}
The raw emotional scores indicate a seemingly neutral state, which traditional systems would classify as emotionally stable. However, the CHS leverages the high contextual drain value (0.9) to calculate a significant $X_{drain}$, resulting in critically low stability despite minimal emotional intensity.

This configuration effectively models burnout or emotional exhaustion states that are not captured by discrete emotion labels but represent genuine psychological distress.

\textbf{Generated Token}: \texttt{qx0AAAAZzAAAZgA=}

\textbf{Key Finding}: CHS successfully represents forms of psychological distress (burnout, exhaustion, numbness) that exist outside traditional emotion taxonomies, addressing a significant limitation of categorical emotion models.

\subsection{Case Study 3: Integrated System Performance}
\label{sec:case3}

This case study demonstrates the full CHS framework processing a complex emotional scenario involving multiple conflicting emotions, high contextual stress, and significant psychological load.

\subsubsection{Input and Raw Processing}
\textbf{Input Text}: ``I was so excited about getting this promotion at work - I've been working toward it for three years. But now that I have it, I feel completely overwhelmed. My new team looks at me like I don't belong here, and honestly, sometimes I agree with them. I'm proud of what I've accomplished, but I'm also terrified I'm going to fail and prove everyone right. I miss the simplicity of my old job, even though I know I should be grateful for this opportunity. I can't sleep at night thinking about all the ways I could mess this up.''

\subsubsection{CHS Output Analysis}
\textbf{Final EmotionalState}:
\begin{itemize}
    \item \textbf{Coordinates}: $(0.34, -0.71)$
    \item \textbf{Dominant Emotion}: Fear
    \item \textbf{Stability}: $S = 0.0$ (Complete Breakdown/Critical State)
    \item \textbf{Generated Token}: \texttt{qyUAAMwAgAAzmQA=}
    \item \textbf{Processing Time}: 1.7 seconds
\end{itemize}

\subsubsection{Multi-Drain Analysis}
The system simultaneously processes three distinct drain mechanisms:

\begin{enumerate}
    \item \textbf{Emotional Load Drain ($E_{drain} = 1.1$)}: High intensities across Fear (0.8), Sadness (0.6), Joy (0.5), and Pride (0.2) create substantial emotional load.
    
    \item \textbf{Conflict Drain ($C_{drain} = 0.5$)}: Opposing emotions (Pride vs. Fear/Guilt, Joy vs. Sadness) generate psychological tension requiring energy to maintain.
    
    \item \textbf{Contextual Drain ($X_{drain} = 0.8$)}: Environmental stressors including work pressure, social pressure, insomnia, and major life changes contribute to overall instability.
\end{enumerate}

The integration of these three drain sources results in zero stability, correctly identifying a critical psychological state requiring intervention.

\textbf{Key Finding}: The CHS framework successfully synthesizes complex, multi-faceted emotional experiences into a single, interpretable psychological profile that captures both the intensity and the inherent instability of the emotional state.

\subsection{Discussion}
\label{sec:discussion}

These case studies demonstrate three core advantages of the CHS framework over traditional emotion detection systems:

\begin{enumerate}
    \item \textbf{Conflict Modeling}: Unlike systems that treat emotions as independent variables, CHS recognizes and quantifies the psychological cost of emotional conflicts.
    
    \item \textbf{Contextual Integration}: The framework captures forms of distress that exist outside discrete emotion categories, such as burnout and emotional numbness.
    
    \item \textbf{Holistic Assessment}: By combining multiple drain mechanisms, CHS provides a comprehensive stability metric that reflects the true complexity of human emotional experience.
\end{enumerate}

The tokenization system enables efficient storage and comparison of emotional states, while the coordinate representation facilitates visualization and analysis of emotional trajectories over time.
\section{Limitations \& Future Work}

\subsection{Current Theoretical Limitations}
This work presents a foundational mathematical framework without empirical validation. The proposed 8-core emotion mapping to 2D space, while theoretically grounded, assumes universal emotional dimensionality that may not capture individual variations in emotional experience. The stability-weighted interpolation algorithm, though psychologically motivated, lacks experimental verification of its accuracy in representing actual emotional transitions.

The current encoding scheme's 8-bit precision may introduce quantization errors, particularly for subtle emotional nuances. Additionally, the fixed coordinate range $[-1, 1]$ may not accommodate extreme emotional states that fall outside this bounded space.

\subsection{Need for Empirical Validation}
The theoretical constructs presented require extensive empirical validation across diverse populations. Future work (ATLAS - Affective Tracking and Learning Analysis System) will focus on:
\begin{itemize}
\item Implementation of real-time emotional tracking algorithms and data structures
\item Performance benchmarking across different hardware architectures
\item Validation through physiological sensor integration and machine learning pipelines
\item Comparative analysis of prediction accuracy against established emotional models
\item Code optimization and scalability testing for production deployment
\end{itemize}

\subsection{Cross-Cultural Considerations}
The current framework assumes universal emotional coordinate mappings, which may not hold across different cultural contexts. Emotional expression, interpretation, and stability patterns vary significantly between cultures. Future research must examine:
\begin{itemize}
\item Cultural variations in emotional coordinate positioning
\item Different stability baselines across populations
\item Language-specific emotional categorizations affecting the 8-core emotion model
\item Adaptation mechanisms for culturally-specific emotional frameworks
\end{itemize}

\subsection{Scalability Questions}
Real-time implementation of this framework raises computational efficiency concerns. The stability-weighted interpolation requires continuous recalculation, potentially limiting deployment in resource-constrained environments. Large-scale emotional tracking systems must address storage and processing requirements for high-frequency emotional state updates.

\subsection{Future Work: 3D Emotional Space}
A promising extension involves migrating from 2D to 3D emotional coordinate systems, where the z-axis explicitly represents stability:
\begin{equation}
\text{EmotionalState}_{3D} = \{(x, y, z) \in \mathbb{R}^3, [I_1, \ldots, I_8] \in [0,1]^8\}
\end{equation}
where $z$ directly encodes stability, eliminating the need for separate stability interpolation algorithms. This would enable more intuitive visualization of emotional trajectories and potentially capture temporal emotional dynamics more naturally.

Additional future directions include extending to continuous emotion detection, real-time adaptation algorithms, and integration with emerging affective computing platforms.

\section{Implementation and Reproducibility}
To support reproducibility and facilitate further research, we have implemented the complete Coordinate Heart System framework in open-source software. The implementation includes all mathematical operations described in this paper, visualization capabilities, and example applications demonstrating the system's practical utility.

The source code is publicly available at \texttt{https://github.com/omar-aldesi/coordinate-heart-system} and includes comprehensive documentation for researchers interested in applying or extending this framework.

\section{Conclusion}
This paper presents a novel mathematical framework for representing and manipulating emotional states in computational systems. Our key theoretical contributions establish a foundation for precise emotional modeling that bridges psychological theory with practical implementation requirements.
\subsection{Key Theoretical Contributions}
We have introduced three fundamental innovations in computational emotion representation:
\textbf{Unified Emotional State Representation:} The integration of 2D coordinate mapping, 8-core emotion intensities, and stability metrics into a single mathematical construct provides unprecedented granularity in emotional state description. This unified approach captures both the dimensional and categorical aspects of emotion theory within a computationally tractable framework.
\textbf{Stability-Weighted Interpolation Algorithm:} Our novel interpolation method addresses a critical gap in emotional state transitions by incorporating psychological principles of emotional inertia. This algorithm prevents unrealistic emotional volatility while maintaining system responsiveness, offering a more psychologically plausible model of emotional dynamics.
\textbf{Efficient Encoding Scheme:} The 8-bit tokenization system enables practical deployment across diverse computational platforms while preserving essential emotional information. This encoding bridges the gap between high-precision emotional modeling and resource-constrained implementation environments.
\subsection{Mathematical Framework Established}
The comprehensive mathematical foundation developed in this work provides:
\begin{itemize}
\item Rigorous definitions for emotional state spaces and transformations
\item Validated interpolation algorithms with psychological grounding
\item Scalable encoding mechanisms suitable for real-time applications
\item Clear pathways for extending to higher-dimensional emotional representations
\end{itemize}
These mathematical constructs offer researchers and developers a standardized approach to emotional computation, facilitating reproducible research and cross-platform compatibility.
\subsection{Foundation for Practical Applications}
The theoretical framework presented here establishes the groundwork for transformative applications in affective computing. The mathematical rigor ensures reliable implementation while the psychological foundation guarantees meaningful emotional interpretation.
Our work directly enables the development of advanced emotional tracking systems, adaptive user interfaces, and personalized affective computing platforms. The stability-weighted interpolation particularly opens new possibilities for smooth emotional state management in real-time applications.
The upcoming ATLAS system will demonstrate the practical potential of this framework through comprehensive implementation and validation studies. By establishing this theoretical foundation, we enable the next generation of emotionally-aware computational systems that can understand, predict, and respond to human emotional states with unprecedented accuracy and sensitivity.
This mathematical framework represents a crucial step toward truly empathetic artificial intelligence systems capable of nuanced emotional understanding and appropriate affective responses.

\bibliographystyle{plain} 
\bibliography{references}

\begin{thebibliography}{1}

\bibitem{barrett2006solving}
Lisa~Feldman Barrett.
\newblock Solving the emotion paradox: Categorization and the experience of emotion.
\newblock {\em Personality and Social Psychology Review}, 10(1):20--46, 2006.

\bibitem{ekman1992argument}
Paul Ekman.
\newblock An argument for basic emotions.
\newblock {\em Cognition and Emotion}, 6(3-4):169--200, 1992.

\bibitem{fontaine2007world}
Johnny R.~J. Fontaine, Klaus~R. Scherer, Etienne~B. Roesch, and Phoebe~C. Ellsworth.
\newblock The world of emotions is not two-dimensional.
\newblock {\em Psychological Science}, 18(12):1050--1057, 2007.

\bibitem{kim2013emotional}
Evgeny Kim and Roman Klinger.
\newblock Who feels what and why? annotation of a literature corpus with semantic roles of emotions.
\newblock {\em Proceedings of the 27th International Conference on Computational Linguistics}, pages 1345--1359, 2013.

\bibitem{lovheim2012new}
Hugo Lövheim.
\newblock A new three-dimensional model for emotions and monoamine neurotransmitters.
\newblock {\em Medical Hypotheses}, 78(2):341--348, 2012.

\bibitem{mehrabian1996pleasure}
Albert Mehrabian.
\newblock {\em Pleasure-Arousal-Dominance: A General Framework for Describing and Measuring Individual Differences in Temperament}, volume~14.
\newblock Current Psychology, 1996.

\bibitem{plutchik2001nature}
Robert Plutchik.
\newblock The nature of emotions: Human emotions have deep evolutionary roots.
\newblock {\em American Scientist}, 89(4):344--350, 2001.

\bibitem{posner2005circumplex}
Jonathan Posner, James~A. Russell, and Bradley~S. Peterson.
\newblock The circumplex model of affect: An integrative approach to affective neuroscience, cognitive development, and psychopathology.
\newblock {\em Development and Psychopathology}, 17(3):715--734, 2005.

\bibitem{russell1980circumplex}
James~A. Russell.
\newblock A circumplex model of affect.
\newblock {\em Journal of Personality and Social Psychology}, 39(6):1161--1178, 1980.

\end{thebibliography}
\end{document}